\documentclass[10pt,twocolumn,letterpaper]{article}

\usepackage{wacv}      

\usepackage{graphicx}
\usepackage{graphicx}
\usepackage{amsthm}
\newtheorem{remark}{Remark}

\usepackage{amsmath}
\usepackage{amssymb}
\usepackage{booktabs}
\usepackage[square,numbers,sort&compress]{natbib}
\usepackage{stfloats}
\usepackage{caption} 
\usepackage{multirow}
\usepackage[breaklinks,hidelinks]{hyperref}

\usepackage[capitalize]{cleveref}
\crefname{section}{Sec.}{Secs.}
\Crefname{section}{Section}{Sections}
\Crefname{table}{Table}{Tables}
\crefname{table}{Tab.}{Tabs.}

\def\wacvPaperID{2993} 
\def\confName{WACV}
\def\confYear{2027}

\begin{document}

\title{Taming Neural Fields with a Frequency-Lipschitz Curriculum for Sparse-View CT Reconstruction}

\author{%
{\normalsize Songtao Hu\textsuperscript{1,*}\quad
Jingrui Wu\textsuperscript{2,*}\quad
Hong Ding\textsuperscript{3}\quad
Fanhao Zhu\textsuperscript{4}\quad
Yunhao Liu\textsuperscript{5}\quad
Suyang Xi\textsuperscript{6,\textdagger}}\\[4pt]
{\small \textsuperscript{1}Xiamen University\quad
\textsuperscript{2}Rice University}\\
{\small \textsuperscript{3}University of Illinois at Chicago\quad
\textsuperscript{4}Reading University}\\
{\small \textsuperscript{5}Hong Kong Polytechnic University\quad
\textsuperscript{6}Emory University}\\[3pt]
{\small \textsuperscript{*}Equal contribution.\quad \textsuperscript{\textdagger}Corresponding author: \texttt{sxi5@emory.edu}}
}

\twocolumn[{
  \maketitle
  \vspace{-1em}
  \centering
  \includegraphics[width=0.8\textwidth, height=0.31\textheight, keepaspectratio]{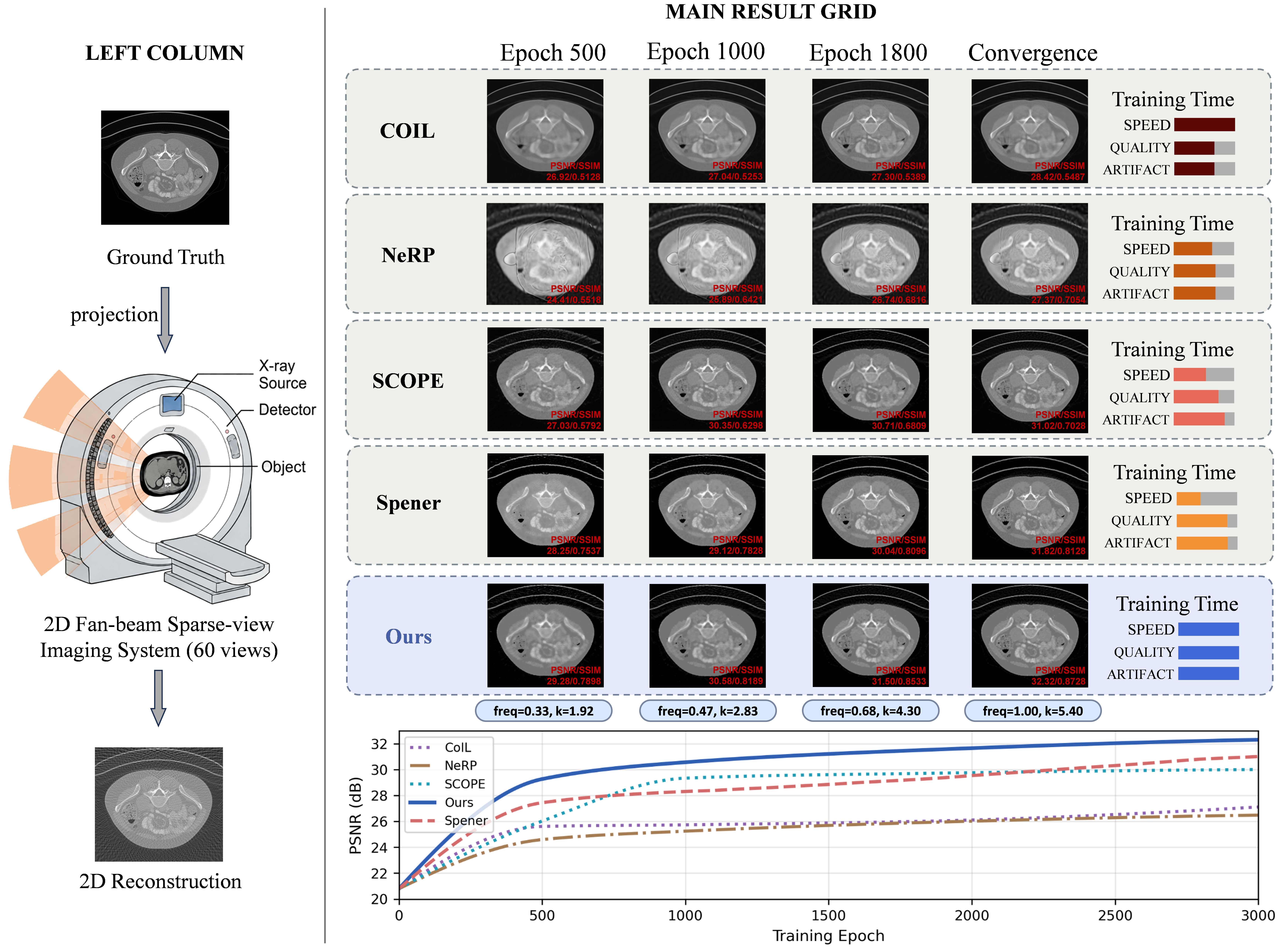}
  \captionof{figure}{Sparse-view CT reconstruction on a single AAPM slice with 60 
views. A comparative analysis with CoIL~\cite{sun2021coil}, 
NeRP~\cite{shen2022nerp}, SCOPE~\cite{wu2023self}, and 
Spener~\cite{tian2025unsupervised} highlights the superior 
convergence speed, reconstruction quality, and artifact suppression 
of the proposed method without iterative denoising.}
  \label{fig:framework}
  \vspace{1.0em}
}]

\begin{abstract}
Sparse-view computed tomography (SVCT) reduces radiation dose by acquiring fewer projection views, but the resulting under-determined inverse problem requires careful regularization to prevent reconstruction algorithms from fitting streak artifacts and hallucinated structures. Existing implicit neural representation (INR) methods rely on implicit spectral bias or iterative external denoising, yet they provide limited control over when high-frequency capacity is exposed during optimization. To address this limitation, we propose \textbf{FLIC} (Frequency-Lipschitz INR Curriculum), a capacity-controlled self-prior INR framework that makes the network's representational capacity explicit and schedulable. The core of our method is a joint frequency-Lipschitz curriculum that couples random Fourier feature encoding with spectrally normalized MLP decoding under non-uniform Lipschitz budgets, guiding reconstruction along a coarse-to-fine trajectory from stable global anatomy to fine details. The terminal capacity is automatically calibrated from FBP gradient statistics, enabling scan-adaptive regularization without per-scan manual tuning of the terminal Lipschitz budget. We further introduce a prior-guided adaptive ray sampling strategy that concentrates integration samples near high-gradient anatomical boundaries, improving forward-model fidelity. The entire framework operates in a single training stage without iterative denoising, enabling efficient per-image reconstruction. Extensive experiments on four public CT datasets under noise-free and low-dose settings show that our method achieves state-of-the-art unsupervised reconstruction quality with consistent performance across diverse anatomical structures.
\end{abstract}

\section{Introduction}
\label{sec:intro}

X-ray computed tomography (CT) is a cornerstone of clinical diagnosis, providing non-invasive visualization of internal anatomy with high spatial resolution \cite{kak2001principles}. However, its reliance on ionizing radiation raises cumulative safety concerns, motivating dose-reduction protocols under the As Low As Reasonably Achievable (ALARA) principle \cite{brenner2007computed, slovis2002alara}. Sparse-view CT (SVCT) offers a direct and hardware-compatible route to dose reduction by acquiring fewer projection views during scanning \cite{sidky2008image}. This reduction, however, leaves the measured sinogram insufficient to fully constrain the underlying attenuation field, making SVCT a severely under-determined inverse problem with a large null space \cite{frikel2013characterization}. As angular sampling becomes sparse, fine anatomical structures are weakly supported by the measurements, and reconstruction algorithms without adequate regularization can explain the missing information by fitting view-dependent streak artifacts or hallucinated structures \cite{wang2020deep}. The central challenge is therefore to design a regularization mechanism that suppresses artifact-consistent solutions while preserving diagnostically relevant anatomical detail.

Existing SVCT reconstruction methods differ mainly in how 
they regularize this under-constrained inverse problem. 
Analytical methods such as FBP~\cite{kak2001principles} are 
efficient but produce severe streak artifacts when views are 
limited~\cite{frikel2013characterization}. Supervised deep 
networks, including FBPConvNet~\cite{jin2017deep} and 
RegFormer~\cite{xia2022transformer}, learn data-driven priors 
from paired reconstructions, but acquiring sufficient paired 
data is costly and the learned priors degrade under shifts in 
anatomy or acquisition protocol~\cite{antun2020instabilities, 
han2018framing}. In contrast, INR-based unsupervised methods 
optimize coordinate-based neural representations directly from 
the measured projections without large-scale training data. 
NeRP~\cite{shen2022nerp} incorporates information from a prior 
image into an INR and optimizes it using the physics of the 
sparsely sampled measurements, while CoIL~\cite{sun2021coil} 
and SCOPE~\cite{wu2023self} exploit neural representations 
and their spectral properties for implicit regularization. 
Spener~\cite{tian2025unsupervised} further introduces an 
image-domain self-prior through a CNN encoder and refines the 
reconstruction via PnP-HQS iterations with BM3D 
denoising~\cite{dabov2007image}. Despite their data-free 
advantage, these methods provide limited explicit control over 
network capacity during optimization or rely on computationally 
expensive external regularization.

This limitation reveals a key tension in sparse-view CT reconstruction: not all recoverable signals are equally reliable under incomplete angular measurements. Low-frequency anatomical structures are relatively well supported by sparse projections and should be established early, whereas fine-scale details remain weakly constrained until the global anatomy has been reliably recovered. When an INR is allowed to express high-frequency content too early, optimization may explain the measured sinogram by fitting view-dependent streak patterns that are absent from the true anatomy. Conversely, overly restrictive representations suppress diagnostically relevant details. The key challenge is therefore not merely to impose stronger regularization, but to explicitly control when and how the network's representational capacity is released during training, so that reconstruction follows a coarse-to-fine trajectory from stable global structure to reliable fine detail.

To address this challenge, we propose \textbf{FLIC}, which departs from conventional solution-level regularization by treating the network's representational capacity itself as the regularization target, controlling not how much the output is penalized but when the network is allowed to become expressive. Concretely, we replace hash-based coordinate 
encodings~\cite{muller2022instant} with random Fourier features 
(RFF)~\cite{tancik2020fourier}, whose frequency spectrum and 
Lipschitz constant admit closed-form expressions, and constrain 
the subsequent MLP via spectral 
normalization~\cite{miyato2018spectral} with non-uniform 
per-layer Lipschitz budgets~\cite{mcginnis2026beyond}. These 
two controls are coupled into a joint frequency-Lipschitz 
curriculum that starts from a narrow frequency window and a 
conservative Lipschitz budget to establish stable anatomical 
structure, and gradually releases capacity to resolve fine 
details. The terminal budget is automatically calibrated from 
the gradient statistics of the FBP reconstruction, adapting 
the curriculum to the signal complexity of each scan.

To improve forward-model accuracy, we further introduce a prior-guided adaptive ray sampling strategy that allocates integration samples to high-gradient regions estimated from the FBP reconstruction. Unlike plug-and-play self-prior methods requiring repeated external denoising, our method follows a single-stage coarse-to-fine trajectory within the INR. Experiments on AAPM \cite{4yqw-2364-22}, COVID-19 \cite{DVN/6ACUZJ_2021}, CQ500 \cite{chilamkurthy2018development}, and ToothFairy \cite{bolelli2024segmenting} show competitive or superior reconstruction quality across sparse-view and dose settings with reduced per-image reconstruction time. Our main contributions are:

\begin{itemize}
\item We analyze the multiplicative coupling between frequency exposure and Lipschitz-constrained capacity in INR-based SVCT. Based on this analysis, we propose a joint frequency-Lipschitz curriculum that progressively releases capacity from stable low-frequency structures to fine anatomical details, enabling single-stage coarse-to-fine reconstruction without external iterative denoising.

\item We introduce a scan-adaptive budget calibration strategy that derives the terminal Lipschitz budget from FBP gradient statistics, adapting regularization strength to scan-specific structural complexity without per-region manual tuning.

\item We develop a prior-guided adaptive ray sampling strategy that allocates samples according to FBP-derived anatomical gradients, improving forward-model fidelity near high-gradient boundaries under sparse-view acquisition.
\end{itemize}

\section{Related Works}

Related medical vision-language research spans report-grounded segmentation, visual question answering, radiographic reasoning, and pathology pre-training~\cite{xi2026grounding,xi2026medlvr,xi2026radiographic,yang2025enhancing}.
\label{sec:related}

\subsection{Implicit Neural Representations for Sparse-View CT}
\label{sec:related_inr}

Implicit neural representations (INRs) parameterize the continuous attenuation field as a coordinate-conditioned neural network optimized against the measured sinogram, avoiding external training data~\cite{xie2022neural}. Existing INR-based SVCT methods mainly follow two regularization paradigms. The first relies on intrinsic neural priors without external iterative denoising. IntraTomo~\cite{zang2021intratomo} combines Fourier-feature-encoded MLPs with iterative geometry refinement and image priors. CoIL~\cite{sun2021coil} shares parameters across adjacent slices to exploit inter-slice consistency. NeRP~\cite{shen2022nerp} embeds prior-image information into the neural representation and optimizes it using measured projections. SCOPE~\cite{wu2023self} separates reconstruction into low- and high-frequency stages, leveraging the spectral bias of MLPs~\cite{rahaman2019spectral}. While effective, these methods offer limited explicit control over when and how representational capacity is released during optimization, which can be insufficient under highly sparse views.

The second paradigm introduces external iterative denoising. Spener~\cite{tian2025unsupervised}
embeds FBP-derived local features into a hash-encoded INR~\cite{muller2022instant}
and alternates data-fidelity optimization with BM3D denoising through a
PnP-HQS scheme~\cite{dabov2007image}. Although this strategy achieves
strong unsupervised performance, the hash encoding has analytically
intractable frequency characteristics, and the multi-round denoising
process substantially increases per-image reconstruction time. In
contrast, our method takes a third path through explicit capacity
control: it retains the self-prior embedding idea, replaces hash encoding
with analytically transparent random Fourier features, and substitutes
external PnP regularization with an intrinsic frequency-Lipschitz
curriculum in a single training stage.

\subsection{Lipschitz Regularization in Neural Networks}
\label{sec:related_lip}

Lipschitz regularization controls the sensitivity of a neural network to input perturbations and provides a principled means of balancing expressiveness and smoothness. Spectral normalization~\cite{miyato2018spectral} constrains layer-wise operator norms, thereby providing an upper bound on the network Lipschitz constant, while alternative methods impose similar constraints through orthogonalized layer designs~\cite{anil2019sorting,araujo2023unified}. For implicit neural representations, a smaller Lipschitz bound promotes smooth outputs but may suppress fine details, whereas a larger bound permits sharper reconstructions at greater risk of fitting noise and measurement artifacts. Lipschitz-constrained INRs have been explored in medical image registration to improve deformation regularity~\cite{wolterink2022implicit,sideri2024sinr}. More recently, the Lipschitz Lens framework~\cite{mcginnis2026beyond} formulates regularization as a task-specific budget-allocation problem, including non-uniform allocation across network components. In contrast, our focus is on training-time capacity scheduling: we explicitly couple the progressive exposure of RFF frequencies with a time-varying Lipschitz budget, thereby controlling when high-frequency representational capacity becomes available during sparse-view CT optimization. This joint frequency-Lipschitz curriculum turns the relationship between encoding bandwidth and Lipschitz-constrained expressiveness into an explicit coarse-to-fine optimization strategy.

\section{Methodology}
\label{sec:method}

\subsection{Preliminaries and Overview}
\label{sec:formulation}

In sparse-view CT reconstruction, the goal is to recover a two-dimensional attenuation field $\mu: \mathbb{R}^2 \to \mathbb{R}_{\geq 0}$ from a set of projection measurements acquired at $N_\theta$ angles, where $N_\theta$ is significantly smaller than the number required for artifact-free analytical reconstruction. Each measurement $y(r)$ corresponds to the line integral of the attenuation field along an X-ray path $r$:
\begin{equation}
    y(r) = \int_r \mu(\mathbf{p}) \, d\mathbf{p},
    \label{eq:forward}
\end{equation}
where $\mathbf{p} \in \mathbb{R}^2$ denotes spatial coordinates along the ray. The collection of all measurements forms the sinogram $\mathbf{y} \in \mathbb{R}^{N_\theta \times N_d}$, with $N_d$ denoting the number of detector elements. When $N_\theta$ is small, the system is severely underdetermined and admits infinitely many solutions consistent with $\mathbf{y}$, necessitating effective regularization to recover a clinically meaningful reconstruction.

Following the self-prior embedding framework of Spener~\cite{tian2025unsupervised}, we parameterize the attenuation field as a neural network $\mathcal{F}_\Phi$ that takes both spatial coordinates and an image-domain prior as input:
\begin{equation}
    \mu(\mathbf{p}) = \mathcal{F}_\Phi(\mathbf{z}, \mathbf{p}),
    \label{eq:inr}
\end{equation}
where $\mathbf{z}$ denotes the FBP reconstruction serving as the prior image, and $\Phi$ denotes the learnable network parameters. The network is trained by minimizing the data consistency loss between predicted and measured sinogram values:
\begin{equation}
    \mathcal{L}_{\mathrm{DC}} = \frac{1}{|\mathcal{R}|} \sum_{r \in \mathcal{R}} \left| y(r) - \hat{y}(r) \right|,
    \label{eq:loss}
\end{equation}
where $\mathcal{R}$ is the set of sampled rays and $\hat{y}(r)$ is the predicted projection obtained by numerically integrating the network output along ray $r$.

\begin{figure*}[t]
  \centering
  \includegraphics[width=0.9\textwidth]{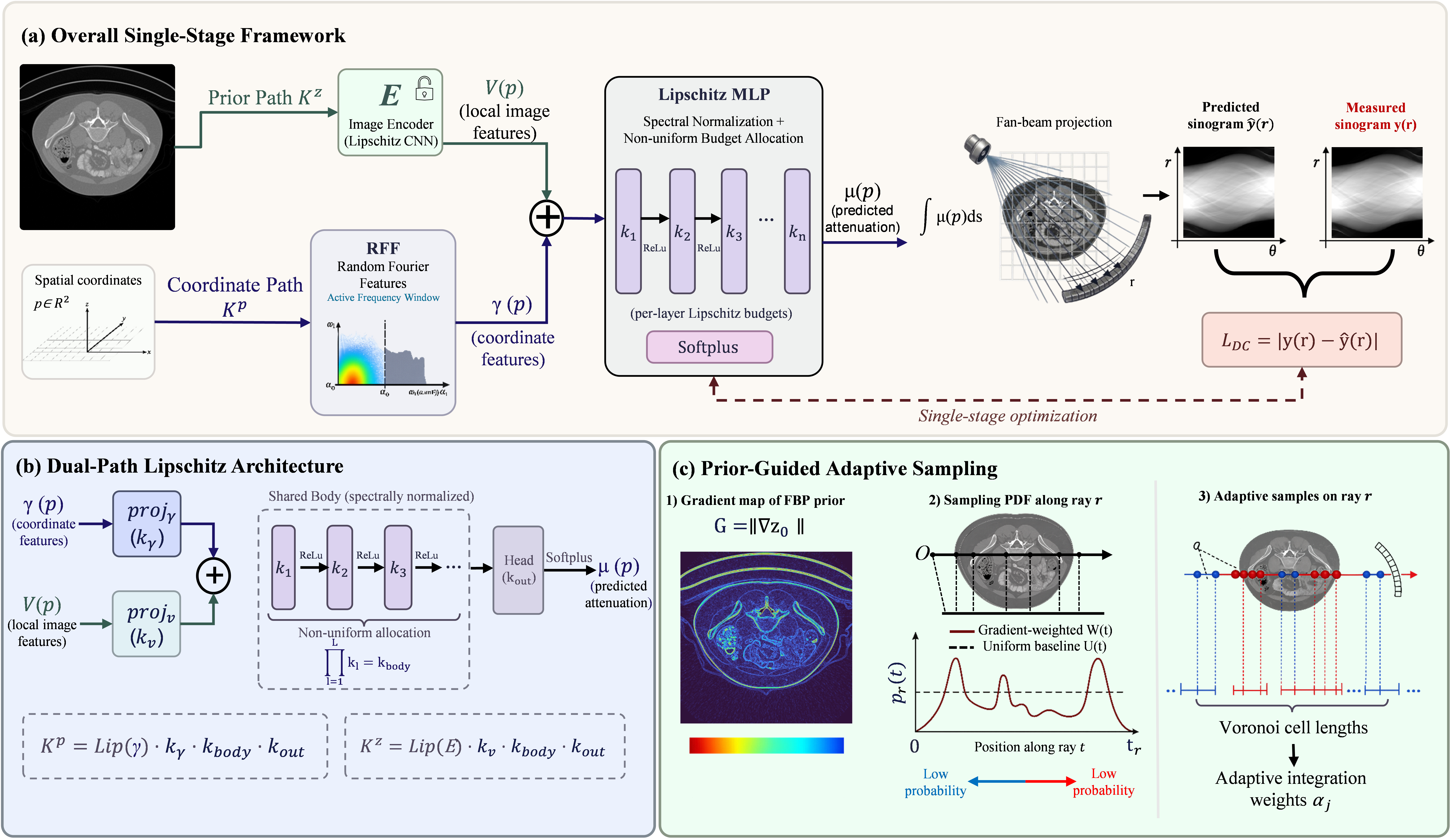}
  \caption{Overview of the proposed framework. (a) Overall single-stage framework. (b) Dual-path Lipschitz architecture. (c) Prior-guided adaptive sampling.}
  \label{fig:overview}
\end{figure*}

Unlike previous methods that regularize the reconstruction 
externally via iterative denoising or implicitly via spectral 
bias, \textbf{FLIC} directly schedules representational capacity 
over the course of optimization.
This is realized through a Lipschitz-constrained architecture 
with dual control paths in Sec.~\ref{sec:architecture}, a 
joint frequency-Lipschitz curriculum for coarse-to-fine 
capacity release in Sec.~\ref{sec:curriculum}, and a 
prior-guided adaptive sampling strategy in 
Sec.~\ref{sec:sampling}. An overview is illustrated in 
Fig.~\ref{fig:overview}.

\subsection{Lipschitz-Constrained Self-Prior Embedding Architecture}
\label{sec:architecture}

The network $\mathcal{F}_\Phi$ consists of three components: a coordinate encoder $\gamma$, an image encoder $\mathcal{E}$, and a Lipschitz-constrained MLP decoder, as illustrated in Fig.~\ref{fig:overview}.

\paragraph{RFF Coordinate Encoder.}
The curriculum in Sec.~\ref{sec:curriculum} derives the MLP 
body budget from the total coordinate-path bound $K^p$, which 
requires the Lipschitz constant of the coordinate encoding to 
be analytically computable. Hash-based 
encodings~\cite{muller2022instant}, whose frequency 
characteristics depend on learned hash collisions, do not 
admit such a bound. We therefore adopt random Fourier features 
(RFF)~\cite{tancik2020fourier}, which satisfy this requirement 
by construction. Spatial coordinates 
$\mathbf{p} \in [-1,1]^2$ are encoded as:
\begin{equation}
    \gamma(\mathbf{p}) = \left[\cos(2\pi \mathbf{B}\mathbf{p}),\; 
    \sin(2\pi \mathbf{B}\mathbf{p})\right] \odot \mathbf{m},
    \label{eq:rff}
\end{equation}
where $\mathbf{B} \in \mathbb{R}^{m \times 2}$ is a fixed 
frequency matrix with rows sampled from 
$\mathcal{N}(\mathbf{0}, \sigma^2 \mathbf{I})$ and sorted by 
magnitude, and $\mathbf{m} \in [0,1]^{2m}$ is a soft mask 
controlling the active frequency window. The sorting enables 
the curriculum to progressively activate frequencies from low 
to high, and the Lipschitz constant admits a closed-form 
expression:
\begin{equation}
    \mathrm{Lip}(\gamma) = 2\pi \sqrt{\lambda_{\max}\!\left(
    \textstyle\sum_{j=1}^{m} m_j^2 \, \mathbf{b}_j 
    \mathbf{b}_j^\top\right)},
    \label{eq:lip_rff}
\end{equation}
where $\mathbf{b}_j$ denotes the $j$-th row of $\mathbf{B}$ and 
$m_j$ the corresponding mask value, making the encoding's 
contribution to $K^p$ fully transparent at every training step.

\paragraph{Image Encoder.}
The image encoder $\mathcal{E}$ extracts local features from the FBP prior image $\mathbf{z}$ through two spectrally normalized convolutional layers with a ReLU activation, producing a feature map from which per-coordinate features $\mathbf{V}(\mathbf{p})$ are obtained via bilinear sampling. The Lipschitz constant of $\mathcal{E}$ is controlled by uniformly distributing a budget $K^z_{\mathrm{enc}}$ across its layers.

\paragraph{Lipschitz MLP Decoder.}
The coordinate features $\gamma(\mathbf{p})$ and image features 
$\mathbf{V}(\mathbf{p})$ are projected into a shared hidden 
space through two spectrally normalized linear layers and summed:
\begin{equation}
    \mathbf{h}_0 = W_\gamma \, \gamma(\mathbf{p}) + W_v \, \mathbf{V}(\mathbf{p}),
    \label{eq:fusion}
\end{equation}
where $W_\gamma$ and $W_v$ have Lipschitz constants $k_\gamma$ 
and $k_v$, respectively. The result is processed by $L$ 
spectrally normalized~\cite{miyato2018spectral} hidden layers 
and a linear head, with a Softplus output activation 
($\beta=10$) to ensure non-negative predictions. Given a total 
body budget $k_{\mathrm{body}}$, the per-layer budgets 
$\{k_\ell\}_{\ell=1}^{L}$ follow an exponential allocation 
strategy~\cite{mcginnis2026beyond}:
\begin{equation}
    k_\ell = k_{\mathrm{body}}^{\,\beta_\ell}, \quad 
    \beta_\ell = \frac{e^{-\ell/L}}{\sum_{i=1}^{L} e^{-i/L}},
    \label{eq:exponential_alloc}
\end{equation}
which assigns larger budgets to earlier layers while 
preserving the global constraint 
$\prod_{\ell=1}^{L} k_\ell = k_{\mathrm{body}}$.

\paragraph{Dual Lipschitz Paths.}
The additive fusion in Eq.~\eqref{eq:fusion} gives rise to two independent Lipschitz chains from input to output. The \emph{coordinate path} bounds how rapidly the predicted attenuation can vary with spatial position:
\begin{equation}
    K^p = \mathrm{Lip}(\gamma) \cdot k_\gamma \cdot \textstyle\prod_{\ell=1}^{L} k_\ell \cdot k_{\mathrm{out}},
    \label{eq:kp}
\end{equation}
A large $K^p$ permits high spatial frequencies in the reconstruction but risks overfitting to streak-pattern artifacts. The \emph{prior path} bounds the network's amplification of noise present in the FBP prior:
\begin{equation}
    K^z = K^z_{\mathrm{enc}} \cdot k_v \cdot \textstyle\prod_{\ell=1}^{L} k_\ell \cdot k_{\mathrm{out}}.
    \label{eq:kz}
\end{equation}
A large $K^z$ propagates streak artifacts from the FBP image into the output. This dual-path decomposition enables independent control: $K^p$ is modulated dynamically via the curriculum (Sec.~\ref{sec:curriculum}), while $K^z$ is held at a fixed conservative value throughout training to suppress prior-image noise.

\subsection{Joint Frequency-Lipschitz Curriculum}
\label{sec:curriculum}

The network's effective spatial bandwidth is determined not by the frequency window or the Lipschitz budget alone, but by their product. From Eqs.~\eqref{eq:lip_rff} and~\eqref{eq:kp}, the coordinate-path bound satisfies:
\begin{equation}
    |\mu(\mathbf{p}_1) - \mu(\mathbf{p}_2)| \leq K^p(t) \cdot \|\mathbf{p}_1 - \mathbf{p}_2\|_2,
    \label{eq:lip_bound}
\end{equation}
where $K^p(t) = \mathrm{Lip}(\gamma_t) \cdot k_\gamma \cdot \prod_{\ell=1}^{L} k_\ell(t) \cdot k_{\mathrm{out}}$ depends multiplicatively on both controls. This coupling is the foundation of our curriculum: scheduling either control in isolation leaves a loophole that the other must close.

\begin{remark}[Capacity coupling]
\label{remark:coupling}
Consider an RFF encoder $\gamma$ with frequency window mask 
$\mathbf{m}$ followed by an $L$-layer MLP $g$ with per-layer 
Lipschitz constants $\{k_\ell\}$. The effective spatial 
bandwidth of the composition $f = g \circ \gamma$ is governed by:
\begin{equation}
    \omega_{\mathrm{eff}} \lesssim \underbrace{\max_{j:\, m_j > 0} 
    \|\mathbf{b}_j\|_2}_{\text{encoding bandwidth}} 
    \cdot \underbrace{\textstyle\prod_{\ell=1}^{L} 
    k_\ell}_{\text{MLP amplification}}.
    \label{eq:freq_bound}
\end{equation}
Controlling only the encoding bandwidth bounds the left 
factor but leaves the right unconstrained, allowing a large 
$\prod k_\ell$ to map low-frequency inputs to high-frequency 
outputs. Conversely, constraining only $\prod k_\ell$ 
suppresses all active frequencies uniformly, preventing 
selective recovery of high-frequency details after 
low-frequency convergence.
\end{remark}

\paragraph{Coupled Schedules.}
The RFF frequency window ratio $r_{\mathrm{freq}}(t)$ grows from $0.2$ to $1.0$, progressively activating higher-frequency basis functions. Simultaneously, the MLP body budget grows linearly:
\begin{equation}
    k_{\mathrm{body}}(t) = k_{\mathrm{body}}^{\min} + (k_{\mathrm{body}}^{\max} - k_{\mathrm{body}}^{\min}) \cdot r(t),
    \label{eq:kbody_schedule}
\end{equation}
where $r(t) = \min(t/(0.8T),\; 1)$ ramps over the first $80\%$ of training and holds at $k_{\mathrm{body}}^{\max}$ for the final $20\%$. Because $\mathrm{Lip}(\gamma_t)$ increases with the opening frequency window while $k_{\mathrm{body}}$ relaxes simultaneously, $K^p(t)$ grows super-linearly as the product of two increasing quantities, releasing capacity multiplicatively so that fine details emerge only on top of already-stabilized low-frequency anatomy.

\paragraph{Signal-Adaptive Budget Calibration.}
Rather than manually selecting $k_{\mathrm{body}}^{\max}$, we 
derive it from the FBP prior image $\mathbf{z}$:
\begin{equation}
    k_{\mathrm{body}}^{\max} = \alpha \cdot \frac{P_{99.9}(\|\nabla 
    \mathbf{z}\|_2)}{\mathrm{median}(\|\nabla 
    \mathbf{z}\|_{2,>0})},
    \label{eq:oracle}
\end{equation}
where $P_{99.9}$ is used instead of the maximum to avoid 
sensitivity to isolated noise spikes in the FBP gradient 
field, and the median over non-zero entries represents the 
typical gradient magnitude in structurally meaningful 
regions. Their ratio captures the dynamic range of the 
gradient field: scans with uniform soft tissue yield a small 
ratio and a tighter budget, while scans with sharp bone--air 
interfaces produce a larger ratio and more permissive budget. 
The scaling factor $\alpha$ is the only free hyperparameter, 
with $k_{\mathrm{body}}^{\min} = 1.0$ corresponding to the 
maximally regularized $1$-Lipschitz baseline. This 
signal-adaptive calibration eliminates per-region 
hyperparameter tuning and enables single-stage training 
without external iterative denoising.

\subsection{Prior-Guided Adaptive Sampling}
\label{sec:sampling}

Standard INR-based CT methods uniformly distribute integration samples along each ray, allocating equal computational resources to homogeneous tissue regions and structurally complex boundaries. We introduce a prior-guided adaptive sampling strategy that concentrates samples toward high-gradient regions identified from the FBP reconstruction.

Given the FBP prior image $\mathbf{z}$, we compute the gradient magnitude field $G(\mathbf{p}) = \|\nabla \mathbf{z}(\mathbf{p})\|_2$ and construct a per-ray sampling distribution as a mixture of a uniform component and a gradient-weighted component:
\begin{equation}
    T \sim \eta \cdot \mathcal{U}[t_n, t_f] + (1-\eta) \cdot \mathcal{W}[t_n, t_f],
    \label{eq:mixture_sampling}
\end{equation}
where $\mathcal{U}$ is the uniform distribution along the ray from near bound $t_n$ to far bound $t_f$, $\mathcal{W}$ is a distribution with density proportional to $G(\mathbf{r}(t)) + \epsilon$, and $\eta \in [0,1]$ controls the mixing weight. The uniform component provides coverage guarantee over the entire ray, while the gradient-weighted component directs additional samples to anatomical boundaries where integration accuracy is most critical. The sampling is performed once before training using the FBP prior, introducing negligible computational overhead. Integration weights are computed using Voronoi-midpoint intervals, providing spacing-aware numerical quadrature for the non-uniform samples along each ray.

\section{Experiments}
\label{sec:experiments}

\subsection{Experimental Setup}
\label{sec:setup}

\paragraph{Datasets and Protocol.}
We evaluate on four public CT datasets: AAPM~\cite{4yqw-2364-22} (abdomen, $512 \times 512$), COVID-19~\cite{DVN/6ACUZJ_2021} (chest), CQ500~\cite{chilamkurthy2018development} (head), and ToothFairy~\cite{bolelli2024segmenting} (dental). Following prior protocols~\cite{sun2021coil, wu2023self, tian2025unsupervised}, we evaluate 10 representative slices per dataset, fixed before comparative evaluation and shared across all methods under identical measurements and acquisition geometry. Additional patient-level robustness results with objectively selected slices are provided in the Appendix. Following~\cite{tian2025unsupervised}, we simulate fan-beam acquisition ($724$-pixel source-to-origin distance, $613$ detectors, $1.3$-pixel spacing) with $720$ dense views, subsampled to $60$ and $90$ sparse views. For noisy measurements, projection-domain Poisson noise is simulated using an incident photon count $I_0$: given a noise-free line integral $y$, the detected photon count is sampled as $I\sim\mathrm{Poisson}(I_0 e^{-y})$, and the noisy projection is computed as $\tilde{y}=-\log(I/I_0)$. We evaluate three dose settings: noise-free, normal dose ($I_0=10^6$), and low dose ($I_0=5\times10^5$). The FBP reconstruction serves as the prior image for all self-prior-based methods. Supervised baselines are trained on the AAPM training set; all other methods operate in an unsupervised per-image manner.

\paragraph{Comparison Methods.}
We compare with FBP~\cite{kak2001principles} as an analytical
baseline; FBPConvNet~\cite{jin2017deep} and
RegFormer~\cite{xia2022transformer} as supervised baselines; and
CoIL~\cite{sun2021coil}, NeRP~\cite{shen2022nerp},
SCOPE~\cite{wu2023self}, and
Spener~\cite{tian2025unsupervised} as unsupervised INR-based methods.
For FBPConvNet and RegFormer, we train separate 60-view and 90-view
models on AAPM using 2{,}400 slices from 8 patients, with 1 patient
reserved for validation. Since these models are not retrained on the
other datasets, they are used as AAPM-trained supervised references
for cross-dataset generalization rather than fully target-domain
competitors. We use the released implementations with default settings
and report PSNR and SSIM~\cite{wang2004image}.

\paragraph{Implementation Details.}
The RFF encoder uses $m = 512$ frequencies with $\sigma = 35.0$. The MLP decoder has $L = 4$ hidden layers of width $128$ with exponential budget allocation. Fixed Lipschitz budgets are $k_\gamma = k_v = k_{\mathrm{out}} = 1.0$ and $K^z_{\mathrm{enc}} = 4.0$; $k_{\mathrm{body}}^{\max}$ is determined by the scan-adaptive budget calibration (Eq.~\eqref{eq:oracle}). The frequency window starts at $r_{\mathrm{freq}} = 0.2$ with $k_{\mathrm{body}}^{\min} = 1.0$. We train for $3{,}000$ epochs using Adam~\cite{kingma2014adam} with learning rate $10^{-3}$. Adaptive sampling uses $\eta = 0.3$. All experiments run on a single NVIDIA RTX PRO 6000 GPU. Under identical data, acquisition geometry, and hardware, the complete reconstruction pipeline takes approximately $7$ minutes per image for Ours versus ${\sim}50$ minutes for Spener; additional complexity and peak-memory comparisons are provided in the Appendix.

\begin{figure*}[t]
\centering
\includegraphics[width=0.9\textwidth]{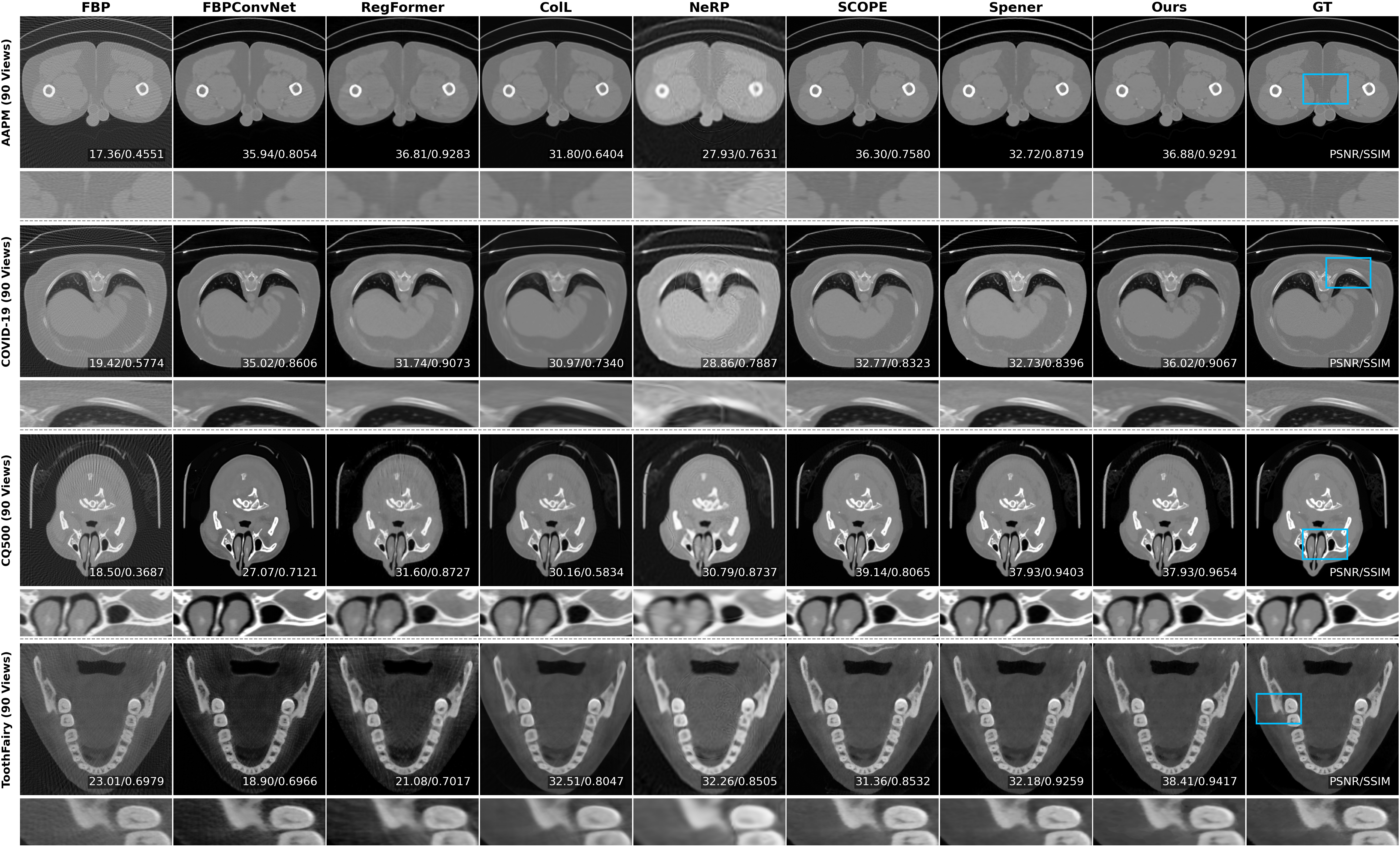}
\caption{Qualitative comparison of reconstructed CT images on
four datasets under the 90-view noise-free setting.
From left to right: FBP, FBPConvNet, RegFormer, CoIL, NeRP, SCOPE,
Spener, Ours, and Ground Truth. Blue boxes highlight regions of
interest, where the proposed method better suppresses streak
artifacts while preserving fine anatomical structures.}
\label{fig:qualitative}
\end{figure*}

\begin{table*}[t]
\centering
\caption{Quantitative comparison on AAPM, COVID-19, CQ500, and ToothFairy datasets under 60-view and 90-view sparse-view settings. Each entry reports PSNR/SSIM. The best and second-best results are highlighted in \textbf{bold} and \underline{underline}, respectively.}
\label{tab:main}
\resizebox{\textwidth}{!}{%
\begin{tabular}{cl cc cc cc cc}
\toprule
\multirow{2}{*}{\textbf{Category}} & \multirow{2}{*}{\textbf{Method}} 
& \multicolumn{2}{c}{\textbf{AAPM}} 
& \multicolumn{2}{c}{\textbf{COVID-19}} 
& \multicolumn{2}{c}{\textbf{CQ500}} 
& \multicolumn{2}{c}{\textbf{ToothFairy}} \\
\cmidrule(lr){3-4} \cmidrule(lr){5-6} \cmidrule(lr){7-8} \cmidrule(lr){9-10}
& & 60 Views & 90 Views & 60 Views & 90 Views & 60 Views & 90 Views & 60 Views & 90 Views \\
\midrule

Analytical 
& FBP 
& 14.38/0.3626 
& 16.94/0.4314 
& 14.15/0.3803 
& 16.29/0.4512 
& 14.52/0.2973 
& 17.52/0.3723 
& 21.03/0.6249 
& 24.50/0.6881 \\

\midrule

\multirow{2}{*}{Supervised} 
& FBPConvNet 
& 33.08/0.8665 
& 35.01/0.8746 
& \underline{28.89}/0.6872 
& 28.82/0.6485 
& 29.06/0.8201 
& 26.65/0.7274 
& 13.01/0.3092 
& 17.10/0.5933 \\

& RegFormer 
& \textbf{34.40}/\underline{0.8748} 
& \underline{35.28}/\textbf{0.9189} 
& 28.58/\textbf{0.7041} 
& \underline{30.26}/\textbf{0.8370} 
& 28.19/0.8096 
& 30.40/0.7969 
& 18.06/0.5656 
& 21.40/0.6811 \\

\midrule

\multirow{5}{*}{Unsupervised} 
& CoIL 
& 28.21/0.5722 
& 30.19/0.6127 
& 25.65/0.5413 
& 27.94/0.5846 
& 24.98/0.4922 
& 27.60/0.5573 
& 28.20/0.7578 
& 32.64/0.8110 \\

& NeRP 
& 27.69/0.7300 
& 28.24/0.7834 
& 27.20/0.6897
& 27.66/0.7393 
& 28.07/0.7609 
& 29.03/0.8475 
& 30.81/0.7896 
& 32.08/0.8542 \\

& SCOPE 
& 32.30/0.7229 
& 34.69/0.7540 
& 27.27/0.6315 
& 29.44/0.6683 
& 32.69/0.8358 
& \underline{35.96}/0.8662 
& 31.27/0.8213 
& \underline{33.73}/0.8528 \\

& Spener 
& 32.13/0.8435 
& 32.78/0.8592 
& 28.68/0.6721 
& 29.73/0.6921 
& \underline{32.99}/\underline{0.8807} 
& 35.87/\underline{0.9329} 
& \underline{31.54}/\underline{0.8764} 
& 32.43/\underline{0.9184} \\

& Ours 
& \underline{33.52}/\textbf{0.8782} 
& \textbf{35.40}/\underline{0.9153} 
& \textbf{29.48}/\underline{0.6918} 
& \textbf{31.14}/\underline{0.7396} 
& \textbf{33.56}/\textbf{0.8974} 
& \textbf{36.28}/\textbf{0.9556} 
& \textbf{34.83}/\textbf{0.8884} 
& \textbf{38.05}/\textbf{0.9377} \\

\bottomrule
\end{tabular}%
}
\end{table*}
\subsection{Comparison with Existing Methods}
\label{sec:comparison}

\paragraph{Noise-Free Setting.}
Table~\ref{tab:main} reports the quantitative results under noise-free acquisition with 60 and 90 views. Among unsupervised methods, our method achieves the highest PSNR and SSIM across all datasets and view settings. Compared with Spener, it improves PSNR by 1.39~dB on AAPM at 60 views, and by 3.29~dB and 5.62~dB on ToothFairy at 60 and 90 views, respectively. These larger gains suggest that the proposed joint frequency-Lipschitz curriculum is particularly effective for preserving sharp bone-air and tooth-tissue boundaries while suppressing streak artifacts. Although the supervised baselines are separately trained on AAPM for each view setting, they degrade on out-of-domain datasets, highlighting the domain-agnostic advantage of per-image unsupervised optimization rather than claiming superiority over target-domain supervised training.

\paragraph{Low-Dose Setting.}
Table~\ref{tab:lowdose} presents results under normal-dose 
($I_0 = 10^6$) and low-dose ($I_0 = 5 \times 10^5$) conditions 
on the AAPM dataset. Under increasing noise levels, existing 
unsupervised INR methods exhibit noticeable performance 
degradation. In contrast, our method maintains robust 
performance, with PSNR dropping by only 0.42~dB from 
noise-free to low-dose at 90 views, compared to 0.60~dB for 
Spener and 1.37~dB for SCOPE. This robustness is attributed 
to the explicit Lipschitz constraint on the prior path $K^z$, 
which limits the network's amplification of noise-corrupted 
FBP features, and the frequency-Lipschitz curriculum, which 
prevents premature fitting of noise-correlated high-frequency 
patterns.


\begin{table*}[t]
\centering
\caption{Quantitative results (PSNR/SSIM) on AAPM dataset under normal and low dose settings.}
\label{tab:lowdose}
\begin{tabular}{llcccc}
\toprule
\multirow{2}{*}{\textbf{Category}} & \multirow{2}{*}{\textbf{Method}} 
& \multicolumn{2}{c}{\textbf{Normal Dose} ($I_0 = 10^6$)} 
& \multicolumn{2}{c}{\textbf{Low Dose} ($I_0 = 5 \times 10^5$)} \\
\cmidrule(lr){3-4} \cmidrule(lr){5-6}
 &  & \textbf{60 Views} & \textbf{90 Views} & \textbf{60 Views} & \textbf{90 Views} \\
\midrule

Analytical 
& FBP 
& 12.26/0.2802 
& 12.31/0.2823 
& 12.23/0.2804 
& 12.26/0.2829 \\

\midrule

\multirow{2}{*}{Supervised} 
& FBPConvNet 
& 32.34/0.8582 
& 34.85/0.7881 
& 31.62/0.8310 
& 34.12/0.7614 \\

& RegFormer 
& \textbf{33.36}/\underline{0.8774} 
& \textbf{35.40}/\underline{0.9127} 
& \underline{32.51}/0.8604 
& \underline{34.55}/0.8811 \\

\midrule

\multirow{5}{*}{Unsupervised} 
& CoIL 
& 27.62/0.5514 
& 29.58/0.5930 
& 27.11/0.5314 
& 29.05/0.5722 \\

& NeRP 
& 27.64/0.7220 
& 28.19/0.7764 
& 27.58/0.7108 
& 28.15/0.7684 \\

& SCOPE 
& 31.63/0.6913 
& 33.83/0.7275 
& 31.25/0.6691 
& 33.32/0.6975 \\

& Spener 
& 31.87/0.8404 
& 32.53/0.8501 
& 31.60/0.8275 
& 32.18/0.8366 \\

& Ours 
& \underline{33.16}/\textbf{0.8771} 
& \underline{35.30}/\textbf{0.9121} 
& \textbf{32.86}/\textbf{0.8692} 
& \textbf{34.98}/\textbf{0.9023} \\

\bottomrule
\end{tabular}
\end{table*}

\paragraph{Qualitative Results.}
Fig.~\ref{fig:qualitative} shows representative 90-view noise-free reconstructions. FBP exhibits severe streak artifacts, while the supervised methods degrade on out-of-distribution datasets. Among unsupervised methods, CoIL retains streaking and blur, SCOPE loses fine details, and Spener shows residual artifacts near sharp transitions. Our method yields the cleanest reconstructions with better-preserved edges across datasets, consistent with Table~\ref{tab:main}.

\begin{table}[t]
\centering
\caption{Ablation study on the AAPM dataset under the noise-free setting. Each entry reports PSNR/SSIM.}
\label{tab:ablation}
\resizebox{\linewidth}{!}{%
\begin{tabular}{lcc}
\toprule
\textbf{Configuration} & \textbf{60 Views} & \textbf{90 Views} \\
\midrule
Baseline (no curriculum)          & 31.84/0.8316 & 33.21/0.8675 \\
+ Frequency curriculum            & 32.46/0.8508 & 34.02/0.8891 \\
+ Lipschitz curriculum            & 32.31/0.8467 & 33.88/0.8846 \\
+ Both (joint curriculum)         & 33.02/0.8669 & 34.73/0.9024 \\
+ Both (reversed direction)       & 32.58/0.8553 & 34.16/0.8917 \\
+ Scan-adaptive budget calibration & 33.27/0.8735 & 35.08/0.9101 \\
+ Adaptive sampling (full)        & \textbf{33.52}/\textbf{0.8782} & \textbf{35.40}/\textbf{0.9153} \\
\bottomrule
\end{tabular}%
}
\end{table}

\subsection{Ablation Study}
\label{sec:ablation}

We conduct ablation experiments on the AAPM dataset to 
validate each proposed component. Table~\ref{tab:ablation} 
reports PSNR/SSIM under 60 and 90 views in the noise-free 
setting.

\begin{figure}[t]
\centering
\includegraphics[width=0.8\columnwidth]{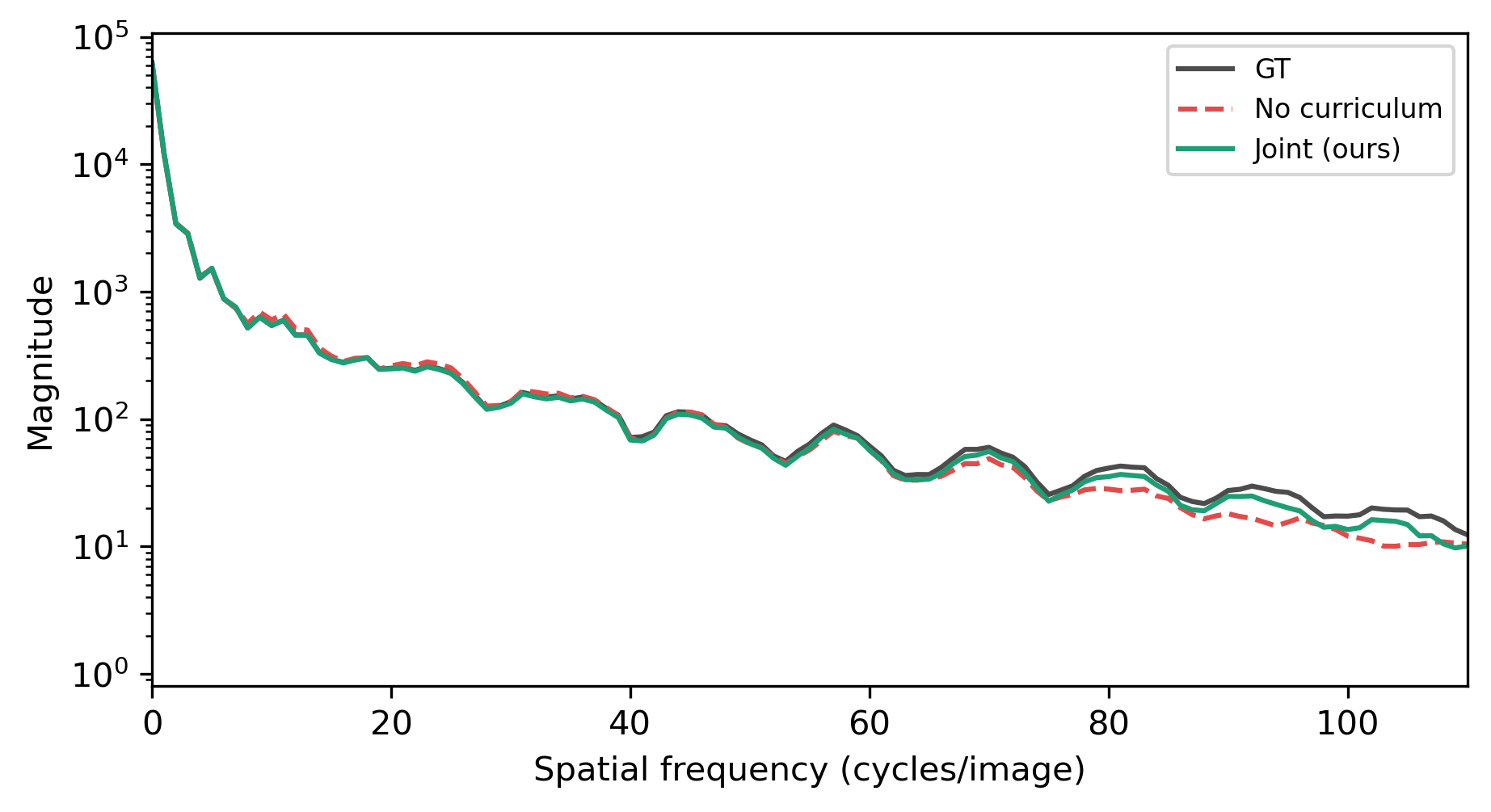}
\caption{Radial frequency spectra on an AAPM test slice (90 views, noise-free). The proposed method better matches the ground-truth spectrum while suppressing excess high-frequency energy.}
\label{fig:spectrum}
\end{figure}

\paragraph{Joint Frequency-Lipschitz Curriculum.}
This experiment evaluates whether frequency exposure and Lipschitz
capacity provide complementary controls for INR-based SVCT. Without
curriculum scheduling, the baseline achieves 31.84/0.8316 and
33.21/0.8675 at 60 and 90 views, respectively. Introducing only the
frequency curriculum improves PSNR by 0.62~dB and 0.81~dB, while using
only the Lipschitz curriculum yields gains of 0.47~dB and 0.67~dB.
Combining both schedules further improves the results to 33.02/0.8669
and 34.73/0.9024, confirming that the two controls are complementary
rather than redundant. Reversing the curriculum direction reduces PSNR
by 0.44~dB and 0.57~dB compared with the joint curriculum, indicating
that the coarse-to-fine ordering is important for stable reconstruction.
This trend is also reflected in Fig.~\ref{fig:spectrum}: the
unconstrained baseline exhibits inflated mid-to-high frequency energy,
whereas the joint curriculum suppresses these spurious components and
more closely follows the ground-truth spectrum.

\paragraph{Scan-Adaptive Budget Calibration and Adaptive Sampling.}
This experiment evaluates whether the remaining components improve
scan-level adaptability and forward-model accuracy. The scan-adaptive
calibration of $k_{\mathrm{body}}^{\max}$ improves the joint curriculum
from 33.02/0.8669 to 33.27/0.8735 at 60 views and from 34.73/0.9024 to
35.08/0.9101 at 90 views, showing that the FBP-derived budget can adapt
the regularization strength using only test-time available information.
Table~\ref{tab:alpha} further shows that PSNR varies by only 0.19~dB
across $\alpha \in [0.3,0.7]$, indicating low sensitivity to this
scaling factor. Finally, prior-guided adaptive sampling provides
additional gains of 0.25~dB and 0.32~dB at 60 and 90 views,
respectively. The full model achieves 33.52/0.8782 at 60 views and
35.40/0.9153 at 90 views.

\begin{table}[t]
\centering
\caption{Sensitivity analysis of the scan-adaptive scaling factor
$\alpha$ on the AAPM dataset (90 views, noise-free).}
\label{tab:alpha}
\resizebox{0.65\linewidth}{!}{%
\begin{tabular}{lccccc}
\toprule
$\alpha$ & 0.1 & 0.3 & 0.5 & 0.7 & 1.0 \\
\midrule
PSNR & 35.12 & \textbf{35.40} & 35.33 & 35.21 & 34.96 \\
SSIM & 0.9106 & \textbf{0.9153} & 0.9138 & 0.9119 & 0.9074 \\
\bottomrule
\end{tabular}%
}
\end{table}

\vspace{-2mm}
\section{Conclusion}
\label{sec:conclusion}

We presented a capacity-controlled INR framework for sparse-view CT reconstruction. By jointly scheduling frequency exposure and Lipschitz-constrained capacity, the proposed curriculum enables coarse-to-fine reconstruction while suppressing premature high-frequency artifact fitting. Scan-adaptive budget calibration and prior-guided sampling further adapt the framework to scan-specific structure without external iterative denoising. Experiments on four public datasets under noise-free and low-dose settings demonstrate competitive reconstruction quality with substantially reduced optimization time. These results support explicit capacity scheduling as an effective alternative to external denoising for INR-based SVCT. Future work will extend the framework to 3D cone-beam CT and learned layer-wise budget allocation.

{\small
\bibliographystyle{ieee_fullname}
\bibliography{egbib}
}
\end{document}